\documentclass[letterpaper]{article} 
\usepackage{aaai2027}  
\usepackage[hyphens]{url}  
\usepackage{graphicx} 
\usepackage{natbib}  
\usepackage{caption} 
\usepackage{algorithm}
\usepackage{algorithmic}
\usepackage{enumitem}
\usepackage{tabularx}
\usepackage{booktabs}
\usepackage{tabularx}
\usepackage{array}
\usepackage{makecell}
\usepackage{amsmath}
\usepackage{newfloat}
\usepackage{listings}
\DeclareCaptionStyle{ruled}{labelfont=normalfont,labelsep=colon,strut=off} 
\floatstyle{ruled}
\newfloat{listing}{tb}{lst}{}
\floatname{listing}{Listing}

\usepackage{booktabs}

\title{RoboFolDeX: A Physical-World Benchmark for Long-Horizon Robotic Manipulation of Deformable Objects}
\author{ Chenhuan Liu\textsuperscript{\rm 1,2}, Yi Xu\textsuperscript{\rm 2}\corresponding, Feng Wu\textsuperscript{\rm 1,2}, Hanyang Wang\textsuperscript{\rm 2}, Wenxiao Kuai\textsuperscript{\rm 2},\\ Weihao Ding\textsuperscript{\rm 2}, Shan Wang\textsuperscript{\rm 2}, Yang Liu\textsuperscript{\rm 3}, Shuyong Gao\textsuperscript{\rm 1}\corresponding, Wenqiang Zhang\textsuperscript{\rm 1}\corresponding }
\affiliations{ \textsuperscript{\rm 1}Fudan University\\ \textsuperscript{\rm 2}AI Research Center, Midea Group (Shanghai) Co., Ltd.\\ \textsuperscript{\rm 3}Carnegie Mellon University }

\begin{document}
\maketitle

\newcommand{\RoboFolDeX}{RoboFolDeX}
\newcommand{\NumTraj}{5,040}
\newcommand{\NumEmbodiments}{4}
\newcommand{\NumTasks}{10}
\newcommand{\NumGarmentTypes}{5}


\begin{abstract}
Embodied AI, including vision-language-action and world-action models, must operate reliably in the \textbf{physical} world. However, models and methods that achieve strong performance in simulation can suffer substantial degradation when deployed on real robots. This gap is particularly severe in long-horizon deformable-object manipulation, where policies must reason over continuously changing states and execute reliable multi-stage bimanual interactions. Existing real-robot benchmarks primarily focus on short-horizon rigid-object manipulation and provide limited coverage of long-horizon deformable tasks.
We introduce \textbf{RoboFolDeX}, a physical-world benchmark built entirely from real-robot data for long-horizon robotic manipulation of \textbf{de}formable objects, with garment \textbf{fold}ing as its primary task. Because collecting real-robot data is expensive, a goal of RoboFolDeX is to study how heterogeneous physical experience can be reused efficiently. We organize the benchmark around four research axes: utilizing human intervention and recovery data collected during policy deployment; transferring data across tasks, including across garment categories and from rigid-object to deformable-object manipulation; reusing data across scenes with changes in lighting, background, and spatial layout; and transferring real-robot data across different embodiments.
RoboFolDeX provides 2,000+ hours of real-robot data spanning 20+ tasks and 10+ robotic embodiments. Building on RoboFolDeX, we establish a fair real-robot evaluation platform for externally submitted policies, with standardized task definitions, held-out physical objects, controlled initializations, and a unified execution protocol. The evaluation platform is publicly accessible at \texttt{https://ai.midea.com/\#/fold-challenge}. We hope that RoboFolDeX will serve as a unified testbed for advancing heterogeneous real-robot data reuse and reliable long-horizon deformable manipulation.
\end{abstract}

\begin{figure*}[t]
    \centering
    \includegraphics[width=0.98\textwidth]{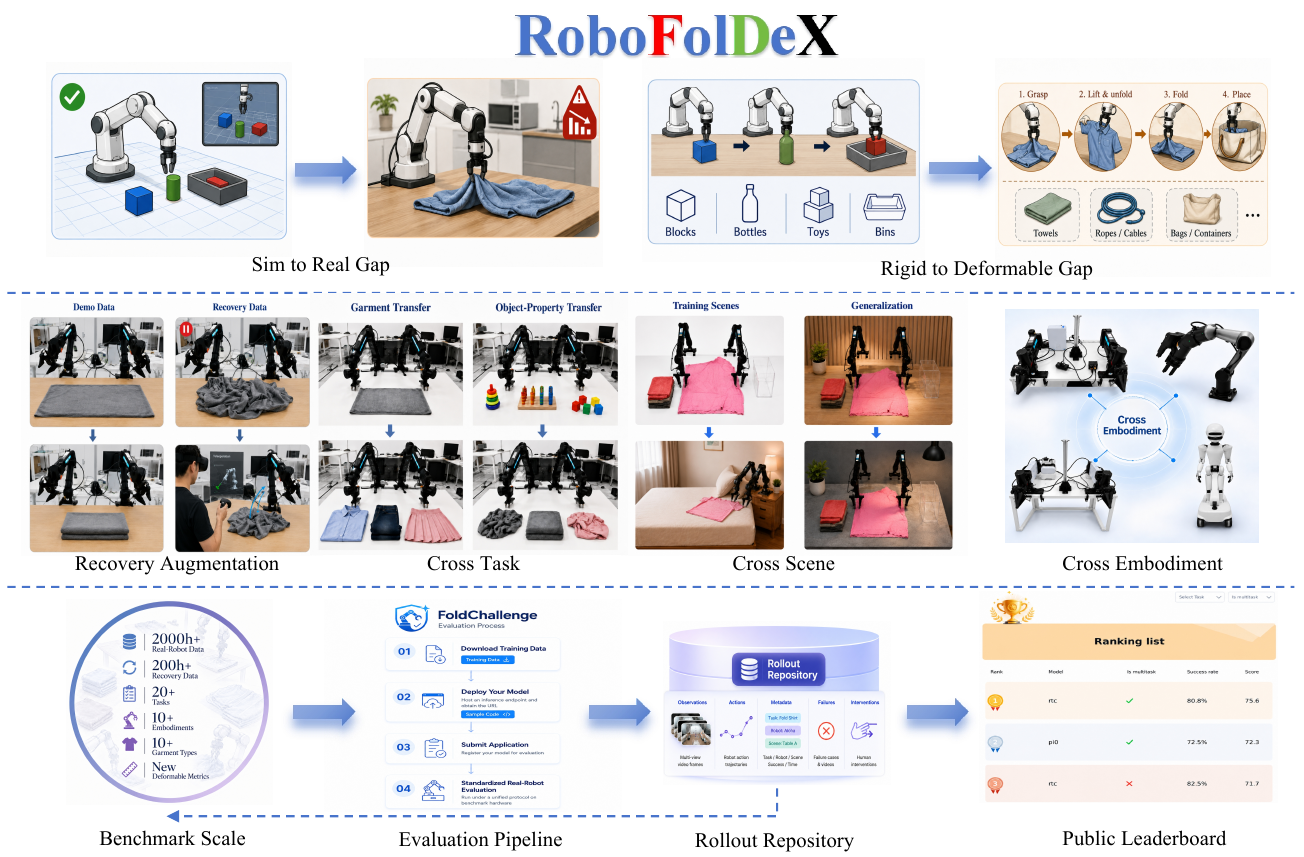}
    \caption{\textbf{\RoboFolDeX{} and the FoldChallenge.}
    \RoboFolDeX{} organizes real-robot data along recovery augmentation, task,
    scene, and embodiment axes. These diagnostic tracks standardize how future
    methods can study robustness and data reuse by holding the target evaluation
    protocol fixed while varying either the available training data or the test
    distribution. FoldChallenge executes externally submitted
    policies under a standardized physical protocol and reports component-wise
    metrics with auditable rollout videos.}
    \label{fig:overview}
\end{figure*}

\section{Introduction}
\label{sec:intro}

The ultimate test of physical AI lies in the real world.
Vision-language-action (VLA) models combine large-scale visual and linguistic
pretraining with robot-trajectory finetuning to predict future action
trajectories from current observations and robot states
~\citep{brohan2023rt2,kim2024openvla,octo2024,black2024pi0}.
Recent world-action models (WAMs) pursue a more ambitious vision:
predicting future visual dynamics may enable embodied models to better
understand the physical world and generate action trajectories that are more
consistent with its underlying dynamics
~\citep{hu2024vpp,ye2026dreamzero,yuan2026fastwam}.
Regardless of the modeling paradigm, however, models and ideas must ultimately
move beyond simulation and curated demonstrations and be validated through
reliable execution on physical robots.

Simulation enables ideas from computer vision, large language models, and
generative modeling to be rapidly transferred to embodied intelligence.
Nevertheless, two fundamental gaps remain.
The first is the \emph{sim-to-real gap}: simulation cannot fully reproduce the
visual, physical, and system-level factors that affect real-robot execution,
including friction, object placement, calibration, latency, sensing noise, and
reset variation.
Consequently, methods that achieve state-of-the-art performance or demonstrate
clear improvements in simulation may degrade substantially or even fail on
physical robots.
The second is the \emph{rigid-to-deformable gap}: methods developed for
rigid-object manipulation often transfer poorly to deformable tasks, where
objects continuously change shape, self-occlude, entangle, and respond
differently after every contact.
This gap is particularly visible in WAMs.
Although future visual prediction provides an appealing mechanism for physical
reasoning, richer visual imagination does not necessarily yield stronger
deformable manipulation.
Recent results from Fast-WAM further show that explicit test-time future
imagination is not always necessary and that a more streamlined,
action-oriented design can perform better on challenging deformable tasks
~\citep{yuan2026fastwam}.
As illustrated in the top row of Figure~\ref{fig:overview}, these two gaps
motivate direct and standardized evaluation in the physical world.

Several real-robot benchmarks have recently emerged to evaluate embodied
policies on physical hardware, including RoboChallenge
~\citep{yakefu2025robochallenge}.
However, two important problems remain.
First, existing real-robot benchmarks primarily emphasize short-horizon
rigid-object manipulation and provide limited coverage of long-horizon
deformable tasks.
This omission is particularly important for household robotics, where garments
and other deformable objects require policies to track continuously changing
states, coordinate contact-rich bimanual actions, preserve progress across
multiple semantic stages, and recover from errors that accumulate throughout
an episode.
Prior work has developed effective systems for smoothing, unfolding, and
folding selected garments
~\citep{avigal2022speedfolding,mo2023foldsformer}, while simulation benchmarks
such as SoftGym and GarmentLab provide scalable deformable-object environments
~\citep{lin2021softgym,lu2024garmentlab}.
Datasets such as Flat'n'Fold further support garment perception and subtask
analysis~\citep{zhuang2025flatnfold}.
Nevertheless, a shared real-robot benchmark for complete, long-horizon
deformable manipulation remains missing.
Second, real-robot evaluation is highly sensitive to data composition, reset
procedures, inference settings, calibration, communication latency, and
engineering implementation.
Even for the same base policy, such as $\pi_0$, incorporating real-time
chunking (RTC) during training can affect deployment performance
~\citep{black2025trainingrtc}.
Without carefully engineered reference implementations and a unified physical
protocol, an apparent improvement may result from an under-optimized baseline,
a favorable data split, or a particular evaluation setting rather than a
meaningful algorithmic advantage.
Model rankings may consequently change when the scene, data composition,
initialization, or system configuration changes, making such comparisons
difficult to reproduce and interpret.

To address these limitations, we introduce \textbf{\RoboFolDeX}, the first
real-robot benchmark focused on long-horizon robotic manipulation of deformable
objects, with garment folding as its primary task family.
\RoboFolDeX{} contains 2,000+ hours of real-robot data spanning 20+ tasks and
10+ robotic embodiments.
Building on \RoboFolDeX, we establish \textbf{FoldChallenge}, a fair and challenging
real-robot evaluation platform with standardized task definitions, held-out
physical objects, controlled initializations, and a unified execution protocol.
We also provide carefully engineered and validated reference implementations,
including a strong $\pi_0$ baseline trained with RTC
~\citep{black2025trainingrtc}, so that future methods can be compared against
competitive real-robot policies rather than weak reproductions.
Because binary success alone cannot characterize the quality of a complete
folding episode, FoldChallenge introduces \emph{FoldScore}, which jointly
considers task success, final-state flatness, and completion time. FoldChallenge is an operational real-robot evaluation platform that was deployed before the submission of this manuscript. It supports externally submitted policies, standardized physical execution, rollout auditing, and leaderboard-based reporting. To preserve double-blind anonymity, identifying information and access links are omitted during review and will be restored in the final version.

Meanwhile, the development of embodied intelligence is increasingly driven by
the pursuit of data scaling.
Large multi-task and multi-robot datasets, interactive data collection, and
cross-embodiment aggregation have greatly expanded the amount and diversity of
available robot experience
~\citep{openx2023,ross2011dagger,liu2023libero,chen2024roviaug}.
However, a key obstacle to scaling real-robot learning may be not only how to
collect more data, but how to reuse the data that have already been collected.
Real-robot data are expensive assets, yet data collected for previous models,
tasks, scenes, or embodiments may not remain effective under a new model or
deployment setting.
Likewise, human intervention, recovery, and policy-rollout data should not be
treated as disposable by-products of a single training cycle.
Understanding whether such experience can be repeatedly incorporated into new
models and settings is essential for establishing meaningful scaling behavior
in physical AI.

Motivated by this perspective, \RoboFolDeX{} organizes heterogeneous real-robot
experience along four research axes.
The \emph{recovery} axis studies how human intervention and recovery data
collected during policy deployment can improve subsequent policies.
The \emph{cross-task} axis studies data transfer across garment categories and
from rigid-object to deformable-object manipulation.
The \emph{cross-scene} axis studies how previously collected data can be reused
when lighting, background, and spatial layout change.
The \emph{cross-embodiment} axis studies how real-robot experience can be
transferred across platforms with different kinematics, sensors, and action
spaces.
For each axis, \RoboFolDeX{} defines standardized data partitions, controlled
comparisons, fixed factors, and reporting requirements, enabling the effects of
heterogeneous data reuse to be measured under consistent physical conditions.

Our contributions are threefold:
\begin{itemize}
    \item We introduce \textbf{\RoboFolDeX}, the first large-scale real-robot dataset
    and benchmark dedicated to long-horizon deformable manipulation, containing
    2,000+ hours of physical robot data across 20+ tasks and 10+ robotic
    embodiments.

    \item We establish \textbf{FoldChallenge}, a publicly released and operational
real-robot evaluation system that supports standardized tasks and controlled
physical execution.

    \item We establish a carefully engineered and highly competitive baseline for
long-horizon folding. Improvements over this baseline under the same training
and physical evaluation protocol would provide strong evidence of meaningful
model advances.
\end{itemize}

\section{Related Work}
\label{sec:related}

\paragraph{Generalist robot policies.}
Vision-language-action (VLA) models map visual observations,
language instructions, and robot states to actions by adapting
pretrained vision-language backbones~\citep{
brohan2023rt2,kim2024openvla,black2024pi0}.
Octo provides an open framework for generalist policy
pretraining~\citep{octo2024}, while $\pi_{0.5}$ studies heterogeneous
co-training for open-world generalization~\citep{
physicalintelligence2025pi05}.
World-action models and predictive policies use future visual
prediction as an auxiliary learning objective~\citep{
hu2024vpp,ye2026dreamzero,yuan2026fastwam}.
As models and training data become more heterogeneous, standardized
evaluation becomes a bottleneck: results obtained under different
robots, task definitions, initial-state distributions, reset procedures,
and scoring rules are difficult to compare directly.

\paragraph{Real-robot benchmarking infrastructure.}
Simulation benchmarks such as CALVIN, LIBERO, and BEHAVIOR-1K
provide scalable evaluation of long-horizon control, transfer, and
household activities~\citep{
mees2022calvin,liu2023libero,li2023behavior1k}.
Complementary systems increasingly support physical-robot evaluation.
RoboChallenge provides remotely accessible platforms and the Table30
benchmark, evaluating generalist policies on 30 real-world tasks using
task success and stage-based progress scores~\citep{
yakefu2025robochallenge}.
RoboDojo combines 42 simulated and 18 physical tasks and provides
RoboDojo-RealEval for remotely hosted evaluation across three robot
embodiments~\citep{chen2026robodojo}.
AutoEval automates success detection, scene reset, and evaluation-job
scheduling on real robot workstations~\citep{zhou2025autoeval},
whereas RoboArena aggregates double-blind pairwise evaluations across
a distributed network of real-robot evaluators~\citep{
atreya2025roboarena}.
VLA-REPLICA specifies a low-cost, locally reproducible physical setup
with standardized in-distribution and out-of-distribution test
scenes~\citep{huang2026vlareplica}.
These systems expand access to real-robot evaluation, but broad
manipulation suites provide limited depth for individual
deformable-object task families.

\paragraph{Deformable-object and garment benchmarks.}
SoftGym standardizes simulated deformable-object manipulation
tasks~\citep{lin2021softgym}, while GarmentLab provides 20 garment and
deformable-object tasks, multiple simulated manipulators, and selected
real-world evaluations~\citep{lu2024garmentlab}.
RGBench focuses on high-fidelity garment simulation and evaluates it
using measured real-world garment dynamics across grasping, folding,
and flinging interactions~\citep{hu2026rgbench}.
The ICRA 2024 Cloth Competition establishes a physical benchmark for
grasp selection in in-air cloth unfolding, with evaluation on a shared
dual-arm platform and 679 unfolding attempts across 34
garments~\citep{degusseme2026clothcompetition}.
Garment folding also appears in broader real-robot suites:
VLA-REPLICA includes towel folding among ten tasks, while
RoboChallenge Table30 includes dishcloth folding~\citep{
huang2026vlareplica,yakefu2025robochallenge}.
These benchmarks provide important physical evaluation precedents,
but emphasize simulation fidelity, specific primitives such as grasp
selection and unfolding, or folding as an isolated task in a
general-purpose suite.
Complementary work studies individual garment-manipulation tasks.
SpeedFolding and FoldsFormer address efficient or sequential garment
folding~\citep{avigal2022speedfolding,mo2023foldsformer}, while
Flat'n'Fold provides synchronized multimodal demonstrations for garment
perception and subtask analysis~\citep{zhuang2025flatnfold}.
In contrast, \RoboFolDeX{} evaluates complete closed-loop, long-horizon
deformable-manipulation episodes, with deep coverage of garment
categories, semantic stages, intermediate and failure states, and
final-state quality under a standardized external physical evaluation
protocol.

\newcommand{\headleft}[1]{%
    \parbox[t]{\linewidth}{%
        \raggedright\bfseries #1\strut
    }%
}

\newcommand{\headcenter}[1]{%
    \parbox[t]{\linewidth}{%
        \centering\bfseries #1\strut
    }%
}

\begin{table*}[t]
    \centering
    \scriptsize
    \setlength{\tabcolsep}{3.2pt}
    \renewcommand{\arraystretch}{1.18}

    \begin{tabularx}{\textwidth}{
        @{}
        >{\raggedright\arraybackslash}p{2.65cm}
        >{\raggedright\arraybackslash}p{2.35cm}
        >{\raggedright\arraybackslash}p{3.45cm}
        >{\centering\arraybackslash}p{1.35cm}
        >{\raggedright\arraybackslash}p{2.35cm}
        >{\raggedright\arraybackslash}X
        @{}
    }
        \toprule

        \headleft{Benchmark} &
        \headleft{Evaluation Setting} &
        \headleft{Deformable-Object Scope} &
        \headcenter{Full LH\\Episode} &
        \headleft{Community Evaluation} &
        \headleft{Reported Signals} \\

        \midrule

        GarmentLab~\citep{lu2024garmentlab} &
        Simulation + real &
        20 garment and deformable-object tasks &
        Selected &
        Local benchmark &
        Task-specific metrics \\[2pt]

        RGBench~\citep{hu2026rgbench} &
        Simulation + real &
        Garment grasping, folding, and flinging &
        No &
        Local benchmark &
        Simulation and task errors \\[2pt]

        ICRA Cloth Competition~\citep{
            degusseme2026clothcompetition} &
        Real &
        In-air cloth-unfolding grasp selection &
        No &
        Live competition &
        Grasp success and final coverage \\[2pt]

        VLA-REPLICA~\citep{huang2026vlareplica} &
        Real &
        General 10-task suite with one towel-folding task &
        Limited &
        Local + leaderboard &
        ID/OOD task success \\[2pt]

        RoboChallenge / Table30~\citep{
            yakefu2025robochallenge} &
        Real &
        General 30-task suite with one dishcloth-folding task &
        Limited &
        Remotely hosted &
        Success and stage progress \\[2pt]

        RoboDojo~\citep{chen2026robodojo} &
        Simulation + real &
        General manipulation; no dedicated deformable track &
        No &
        Remotely hosted &
        Task and capability performance \\[2pt]

        \textbf{\RoboFolDeX{}} &
        \textbf{Real} &
        \textbf{Folding-centered deformable task family} &
        \textbf{Yes} &
        \textbf{External physical evaluation} &
        \textbf{Success, quality, recovery, and efficiency} \\

        \bottomrule
    \end{tabularx}

    \caption{
        \textbf{Comparison with representative manipulation benchmarks.}
        ``Full LH Episode'' indicates evaluation of complete, dependent,
        long-horizon deformable-manipulation episodes.
        ``Limited'' indicates that folding appears as an isolated task
        inside a broader suite rather than a deeply covered task family.
        ``Selected'' indicates that the benchmark includes some
        long-horizon tasks but does not define a benchmark-wide external
        protocol for complete physical deformable-manipulation episodes.
    }
    \label{tab:benchmark_comparison}
\end{table*}

\section{\RoboFolDeX}
\label{sec:RoboFolDeX}

\RoboFolDeX{} is a physical-world benchmark and dataset built entirely from
real-robot data for long-horizon deformable manipulation. It covers continuous
deformation, multi-stage state transitions, contact-rich bimanual coordination,
partial observability, and errors accumulating over long episodes. Garment
folding is the central task because it requires reasoning about garment
structure and continuously changing physical states. The suite also includes
other deformable-object manipulation tasks and a few rigid-object tasks for
studying rigid-to-deformable transfer.

\paragraph{Benchmark scope.}
\RoboFolDeX{} focuses on long-horizon deformable manipulation and the efficient
reuse of real-robot data under heterogeneous data settings. In particular, it
studies how models utilize recovery data and how robot experience transfers
across tasks, scenes, and embodiments.
\RoboFolDeX{} serves three purposes. First, it provides real-robot trajectories for
training and evaluating long-horizon deformable-manipulation policies. Second,
it provides a standardized physical testbed for comparing policies under
consistent real-world conditions. Third, it organizes heterogeneous robot data
along recovery, task, scene, and embodiment to study data-reuse efficiency
and transfer performance.

\begin{figure}[t]
    \centering
    \includegraphics[width=\columnwidth]{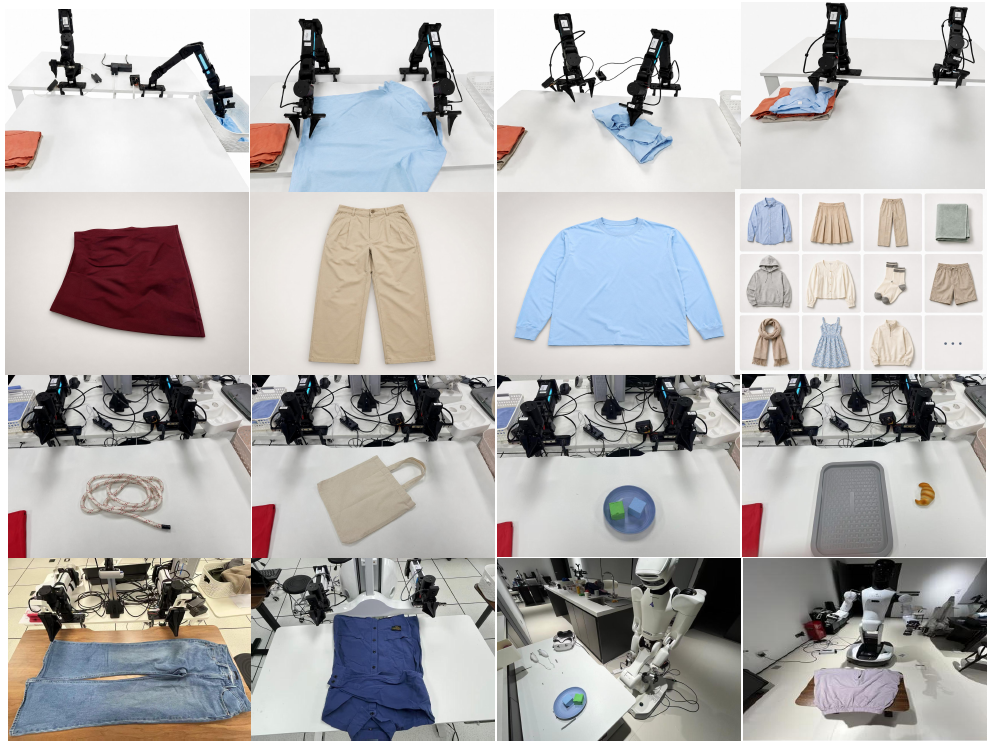}
    \caption{
    \textbf{Overview of \RoboFolDeX{}.}
    Rows show, from top to bottom, the major stages of a long-horizon folding
    task, garment categories, other deformable- and rigid-object manipulation
    tasks, and the robotic embodiments and scenes.
    }
    \label{fig:RoboFolDeX_overview}
\end{figure}

\subsection{Design Principles}
\label{sec:RoboFolDeX_design}

\RoboFolDeX{} follows three design principles. First, it focuses on long-horizon
closed-loop manipulation, where errors in early stages directly affect all
subsequent actions. Second, it emphasizes deformable objects, whose continuously
changing shapes, self-occlusion, and contact dynamics make real-world execution
substantially more challenging than rigid-object manipulation. Third, because
real-robot data are expensive to collect, \RoboFolDeX{} organizes heterogeneous
experience across recovery, task, scene, and embodiment axes to support
systematic data reuse and transfer.

\subsection{Task Suite and Policy Interface}
\label{sec:RoboFolDeX_tasks}

Most \RoboFolDeX{} tasks focus on long-horizon garment folding, with fewer tasks
covering other deformable objects and only a few involving rigid objects.
The rigid-object tasks support studies of joint training and data-efficient
adaptation to deformable manipulation.

\paragraph{Observation and action interface.}
At each inference step, the policy receives synchronized RGB observations from
two wrist-mounted hand--eye cameras and one external workspace camera, together
with a natural-language instruction and the robot proprioceptive state. The
policy predicts a future trajectory of robot end-effector poses.

\paragraph{Long-horizon garment folding.}
A folding episode starts with a garment placed in a basket. The robot retrieves
the garment and places it on the table, where it initially appears in a random
configuration. The robot then flips and flattens the garment, performs multiple
folding operations, and finally places or stacks the folded garment on top of
previously folded garments. These dependent stages form a complete long-horizon
episode from retrieval to final stacking.
Representative tasks, garment categories, stages, robotic embodiments, and
scenes are shown in Figure~\ref{fig:RoboFolDeX_overview}.

\subsection{Data Collection and Organization}
\label{sec:RoboFolDeX_collection}

\RoboFolDeX{} is constructed from 2,000+ hours of heterogeneous real-robot
experience spanning 20+ manipulation tasks and 10+ robotic embodiments. The platforms include Aloha, YAM, Piper,
Astribot, Franka, UR5, a self-developed mobile dual-arm platform,
AgiBot G2, and additional robotic systems. Together, they cover
single-arm, dual-arm, mobile, and humanoid configurations across
tabletop and household environments.

The robotic platforms differ substantially in kinematics, workspace
geometry, camera placement, control interfaces, and action
representations. To support unified training and analysis, the data are
organized using a common schema containing synchronized visual
observations, robot proprioception, timestamped actions,
natural-language task instructions, and episode-level metadata. The
metadata records the task, object category, object instance, scene,
embodiment, operator, data source, completion status, and failure or
termination reason.



\begin{figure*}[t]
    \centering
    \includegraphics[
        width=0.94\textwidth,
        height=0.36\textheight,
        keepaspectratio
    ]{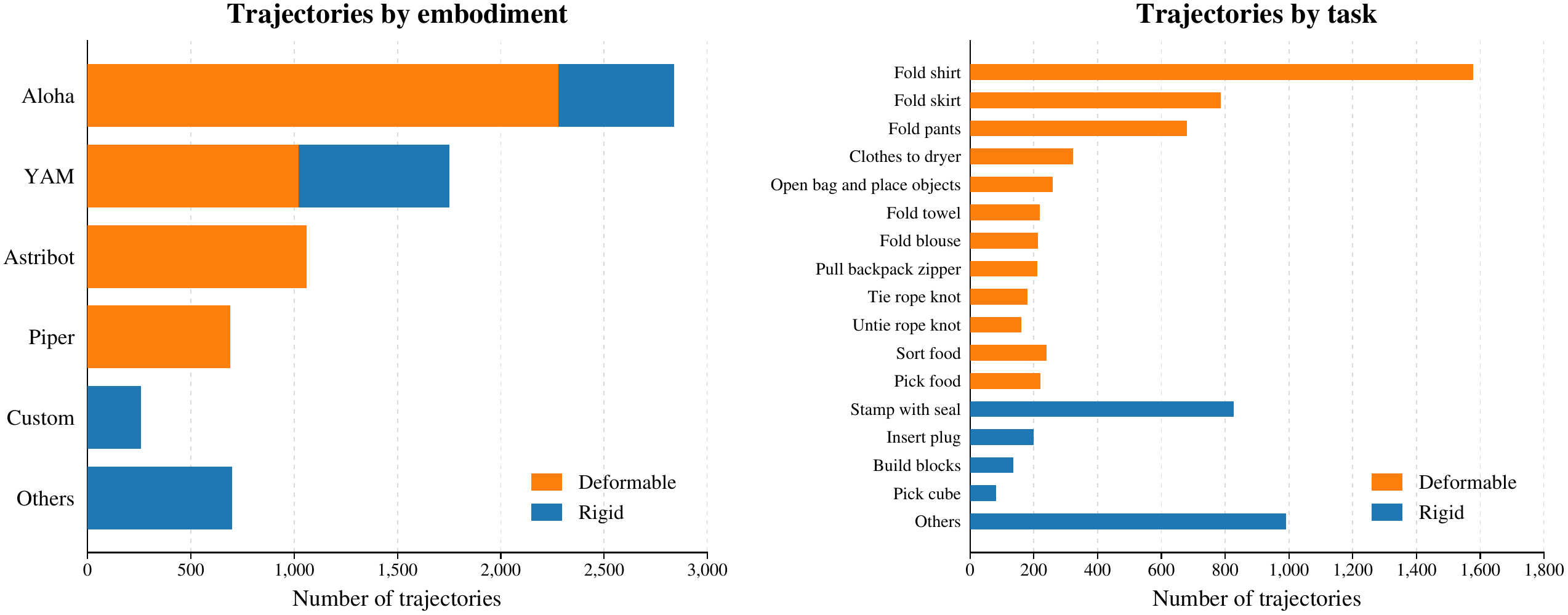}
    \caption{
    \textbf{\RoboFolDeX{} embodiments and data composition.}
    The benchmark contains 2,000+ hours of real-robot experience spanning
    20+ tasks and 10+ robotic embodiments. The
    statistics summarize the distribution of garment-manipulation tasks,
    other deformable-object tasks, rigid-object tasks, and robotic
    embodiments.
    }
    \label{fig:data_statistics}
\end{figure*}



\subsection{Benchmark Tracks}
\label{sec:RoboFolDeX_settings}

\RoboFolDeX{} organizes its real-robot data into four benchmark tracks to evaluate
models and learning methods in recovery-data utilization, multi-task and
cross-task training, multi-embodiment and cross-embodiment training, and
cross-scene transfer and generalization.

\paragraph{Recovery-data utilization.}
A shared policy is first deployed to generate real-robot rollout states, during
which human operators collect intervention and recovery trajectories. The recovery trajectories are combined with
independently collected demonstration data to train and evaluate a separately
initialized policy. This setting evaluates a model's ability to utilize recovery data
and the highest performance achievable under the recovery-augmented
training setup.

\paragraph{Cross-task transfer.}
This track studies whether rigid- and deformable-object data can benefit each
other during joint training. Although garment manipulation is deformable
overall, stages such as garment retrieval and final stacking share action
patterns with rigid-object grasping and placement. The track also evaluates
whether a model trained on a rigid-object task can learn a complete
long-horizon deformable-manipulation task using only a limited amount of
deformable-object data.

\paragraph{Cross-scene transfer and generalization.}
This track changes the robot workspace through variations in spatial layout,
lighting, visual background, and surrounding environment. It evaluates how
much previously collected data remain useful after the scene changes, how
efficiently a policy adapts to the new setting, and how well it generalizes
without additional data.

\paragraph{Cross-embodiment transfer.}
Cross-embodiment transfer is one of the most challenging real-robot settings.
Different robot morphologies have substantially different kinematics, control
interfaces, and action spaces, while modifying the action head can strongly
affect the learned policy. This track studies which task knowledge and action
patterns can be shared across embodiments and whether multi-robot data can
reduce the amount of target-robot data required for learning long-horizon
deformable manipulation.

\subsection{Evaluation Metrics}
\label{sec:RoboFolDeX_metrics}

Binary success rate alone cannot fully characterize performance on complete
garment-folding tasks. Policies with similar success rates may differ
substantially in final-state quality and execution efficiency. We therefore
report task success rate, average completion time, final-state neatness, and
\textbf{FoldScore}.

\paragraph{Final-state neatness.}
We assess the final folded garment using a six-level neatness score from 0 to
5, considering its overall shape, compactness, flatness, structural stability,
and edge alignment. A score of 0 represents a largely unfolded garment,
whereas a score of 5 represents a compact, flat, stable, and well-aligned
result approaching careful human folding quality. The score is determined
solely from the final garment state. The detailed scoring rubric is provided
in the Appendix.

\paragraph{FoldScore.}
FoldScore summarizes final-state quality, task success, and execution
efficiency as
\begin{equation}
\mathrm{FoldScore}
=
35\bar{Q}
+
35R
+
30\max\left(0,1-\frac{t}{T_{\max}}\right),
\end{equation}
where $\bar{Q}\in[0,1]$ is the average normalized final-state quality,
$R\in[0,1]$ is the complete-task success rate, $t$ is the average completion
time over successful trials, and $T_{\max}$ is the maximum episode duration.
The time component is set to zero when no trial succeeds. The mapping from
neatness levels to normalized quality scores is provided in the Appendix.

All policies are evaluated using the same physical tasks, garment instances,
initialization procedure, maximum episode duration, and robot execution
protocol.

\section{Experiments}
\label{sec:experiments}

We conduct preliminary real-robot experiments corresponding to the four
research questions defined by \RoboFolDeX{}: recovery-data utilization,
cross-task transfer, cross-scene transfer, and cross-embodiment transfer.
These experiments provide reference results and comparisons for future
research.
We first establish multi-task reference baselines and compare unified
multi-task training with task-specific training under standard and RTC-based
training configurations.
We then evaluate the effect of augmenting multi-task training with DAgger
recovery data.
Finally, we report preliminary observations on rigid-to-deformable,
cross-scene, and cross-embodiment transfer.
All experiments use the same observation and action interfaces.
Unless otherwise specified, each garment category is evaluated using 30
independent physical trials with randomized garment configurations.
A trial is considered successful only when the policy completes the entire
long-horizon folding sequence without human intervention or unrecoverable
failure.
All models are trained on 16 GPUs with a batch size of 16.
For fair evaluation, all qualitative scores are independently assessed by at
least two evaluators, with the same evaluators used across methods.
A score is recorded only upon agreement.

\subsection{Baselines}
\label{sec:reference_baselines}

We first establish reference baselines for unified multi-task garment folding
and compare them with task-specific policies.
In the multi-task setting, data from different garment categories are mixed
to train a single unified policy.
This setting evaluates whether a shared policy can acquire long-horizon
folding capabilities across multiple garment types.
For comparison, we train a separate task-specific policy for each garment
category using only the corresponding garment data.
This comparison evaluates whether unified multi-task training provides useful
transfer across garment categories or introduces interference between tasks.
All reference models use $\pi_0$ and are trained using expert demonstrations
only.
For multi-task training, we compare standard training with training-time
real-time chunking (RTC).
The task-specific policies are trained with RTC.
Training-time RTC simulates inference delay and conditions the
model on action prefixes that have already been committed for execution,
avoiding inference-time inpainting overhead~\citep{black2025trainingrtc}.

\begin{table}[t]
    \centering
    \small
    \setlength{\tabcolsep}{2.6pt}
    \renewcommand{\arraystretch}{1.08}

    \begin{tabular}{
        @{}
        >{\raggedright\arraybackslash}p{2.25cm}
        >{\centering\arraybackslash}p{1.55cm}
        >{\centering\arraybackslash}p{1.55cm}
        >{\centering\arraybackslash}p{1.55cm}
        @{}
    }
        \toprule
        \textbf{Task / Metric} &
        \shortstack{\textbf{RTC}\\\textbf{Multi}} &
        \shortstack{\textbf{Standard}\\\textbf{Multi}} &
        \shortstack{\textbf{RTC}\\\textbf{Single}} \\
        \midrule

        Shirt SR (\%)        & 93.00 & 90.00 & 100.00 \\
        Shirt FoldScore      & 79.62 & 71.50 & 71.43 \\
        Skirt SR (\%)        & 90.00 & 100.00 & 100.00 \\
        Skirt FoldScore      & 83.38 & 80.00 & 87.57 \\
        Pants SR (\%)        & 63.33 & 50.00 & 80.00 \\
        Pants FoldScore      & 70.52 & 73.33 & 76.23 \\
        Towel SR (\%)        & 76.67 & 50.00 & 50.00 \\
        Towel FoldScore      & 68.87 & 64.40 & 51.54 \\
        \midrule
        \textbf{Average SR (\%)} &
        80.75 & 72.50 & \textbf{82.50} \\
        \textbf{FoldScore} &
        \textbf{75.59} & 72.30 & 71.69 \\
        \bottomrule
    \end{tabular}

    \caption{
        Multi-task and task-specific reference results under the
        FoldChallenge protocol.
        All reference policies use $\pi_0$ and expert demonstrations only.
        RTC denotes real-time chunking during training.
        SR denotes success rate, while FoldScore considers final-state
        quality, complete-task success, and execution time.
    }
    \label{tab:reference_baselines}
\end{table}

Table~\ref{tab:reference_baselines} shows that multi-task and task-specific
training exhibit different advantages under different evaluation metrics.
The RTC task-specific policies achieve the highest average success rate of
$82.50\%$, whereas the RTC multi-task policy obtains the highest average
FoldScore of $75.59$.
This discrepancy illustrates that binary success alone does not completely
characterize the physical quality and efficiency of a folding policy.
Within the multi-task setting, RTC improves the average success rate from
$72.50\%$ to $80.75\%$ and increases the average FoldScore from $72.30$ to
$75.59$.
The improvement is particularly visible on Pants and Towel, which require
longer-horizon state tracking and more substantial manipulation of deformable
configurations.
Task-specific training performs strongly on Shirt, Skirt, and Pants.
In contrast, the RTC multi-task policy substantially outperforms the
task-specific policy on Towel, achieving a success rate of $76.67\%$ compared
with $50.00\%$.
These results indicate that multi-task training can provide useful transfer
for selected garment categories, although its benefits are not uniform across
all tasks.
We use the demo-only RTC multi-task policy as the reference baseline
for evaluating the effect of recovery data.

\paragraph{Supplementary baselines.}
We additionally reproduce $\pi_{0.5}$ and evaluate $\pi_0$ with
inference-time RTC on the Shirt task. The $\pi_{0.5}$ policy is evaluated
from a pre-flattened garment state and therefore covers only the subsequent
folding process, whereas the inference-time RTC baseline is evaluated on the
complete long-horizon task from garment retrieval and flattening to folding
and final placement. The two results are thus reported only as supplementary
references rather than as a direct comparison.
The reproduced $\pi_{0.5}$ policy achieves a success rate of $100.0\%$ with
an average completion time of 41\,s, although its folded results generally
remain in the lowest non-zero final-state quality tier. In comparison,
$\pi_0$ with inference-time RTC achieves a success rate of $42.9\%$ on the
complete task. The main comparison remains restricted to methods evaluated
across all garment categories under the complete evaluation protocol.


\begin{table}[t]
    \centering
    \footnotesize
    \setlength{\tabcolsep}{4pt}
    \renewcommand{\arraystretch}{1.06}
    \begin{tabular}{@{}lccc@{}}
        \toprule
        \textbf{Task} &
        \textbf{Success Rate} &
        \textbf{Average Time} &
        \textbf{FoldScore} \\
        \midrule
        Shirt & 90.00\%  & 2:29 & 76.88 \\
        Skirt & 90.00\%  & 1:54 & 81.25 \\
        Pants & 100.00\% & 1:48 & 85.50 \\
        Towel & 100.00\% & 1:40 & 86.50 \\
        \midrule
        \textbf{Average} &
        \textbf{95.00\%} &
        \textbf{1:58} &
        \textbf{82.53} \\
        \bottomrule
    \end{tabular}
    \caption{
        \textbf{Recovery-augmented multi-task real-robot results.}
A single $\pi_0$ policy with RTC is trained using demonstrations and
recovery data from four garment categories.
    }
    \label{tab:recovery_multitask}
\end{table}

\subsection{Recovery-Data Utilization}
\label{sec:recovery_utilization}

We evaluate the model's ability to utilize recovery data. Starting from the
demonstration-only RTC multi-task baseline in
Table~\ref{tab:reference_baselines}, we add recovery trajectories from Shirt,
Skirt, Pants, and Towel and train a single $\pi_0$ policy with RTC. As shown in Table~\ref{tab:recovery_multitask}, adding recovery data improves
the average success rate from $80.75\%$ to $95.00\%$ and the average
FoldScore from $75.59$ to $82.53$. The improvement is clear on
Pants and Towel, both of which reach a success rate of $100.00\%$.

\subsection{Preliminary Observations on Other Transfer Axes}
\label{sec:preliminary_transfer}

Beyond recovery-data utilization, we conduct preliminary experiments on
cross-task, cross-scene, and cross-embodiment data reuse.
These experiments are intended to verify that the corresponding benchmark
axes represent practically important challenges, rather than to provide
complete solutions to each transfer setting.

For heterogeneous task data, including rigid-object and deformable-object
manipulation, we evaluate both joint training and continued training.
Direct joint training often introduces substantial interference in action
generation, while continued training frequently causes the policy to forget
previously acquired action behaviors.
Although some short manipulation stages can still be completed, the resulting
policies rarely preserve reliable performance over complete long-horizon
deformable-manipulation episodes.
We observe a similar but more severe problem in cross-embodiment training.
We evaluate both joint training across multiple robotic embodiments and
continued training from one embodiment to another.
Differences in robot kinematics, camera configurations, control interfaces,
and action representations lead to severe action-space interference and
catastrophic forgetting.
Under these settings, the policies generally fail to complete the full task
reliably, even when the individual embodiments can be trained successfully
with their own data.

In contrast, policies show stronger robustness to changes in
lighting, background, and moderate workspace layout.
Cross-scene variation is therefore less disruptive than changes in task
dynamics or robotic embodiment in our preliminary evaluations.
These observations indicate that heterogeneous real-robot data cannot be
effectively reused through naive data mixing or continued training alone.
They also motivate the controlled data partitions and evaluation protocols
introduced by \RoboFolDeX{}.
As a benchmark, \RoboFolDeX{} exposes these unresolved challenges and provides a
physical testbed for future methods addressing negative transfer,
catastrophic forgetting, and cross-embodiment action alignment.

\section{Discussion and Limitations}
\label{sec:discussion}
Our comparison is constrained by the time and hardware cost of long-horizon
real-robot evaluation, and additional model evaluations will be continuously
released through FoldChallenge. The four tracks provide dataset partitions,
standardized evaluation settings, and empirical results, rather than exhaustive
methodological studies of recovery, cross-task, cross-scene, and
cross-embodiment transfer. The current benchmark primarily relies on visual
observations and proprioception, without explicit tactile sensing. Future
versions will incorporate tactile feedback to better capture contact states and
local deformations during deformable manipulation.
\section{Conclusion}
We presented \textbf{RoboFolDeX}, the first real-robot benchmark focused on long-horizon deformable manipulation, together with \textbf{FoldChallenge}, a fair and challenging evaluation platform. RoboFolDeX highlights that effectively organizing and reusing heterogeneous real-robot data may be more valuable than blindly scaling data collection. By structuring research across recovery, task, scene, and embodiment axes,
\RoboFolDeX{} provides a foundation for reliable embodied AI and physical-world
robotic manipulation.
\paragraph{Use of Generative AI.}
Generative AI tools were used for language editing and visual drafting; all
technical content, results, and final figures were verified by the authors.
\newpage
\bibliography{aaai2027}

\end{document}